\documentclass[journal,10pt,onecolumn,draftclsnofoot,]{./IEEEtran}

 \pdfoutput=1

\usepackage[T1]{fontenc}

\usepackage{cite}
  \usepackage{graphicx}
  \graphicspath{{./Figures/}}
  \DeclareGraphicsExtensions{.pdf,.jpeg,.png,.tikz}
\usepackage{color}
 \usepackage{multirow}
\usepackage{lipsum}

\usepackage{fixltx2e}

\usepackage[cmex10]{amsmath}
\usepackage{amssymb}
\usepackage{tabularx}
\usepackage{mathtools}
\usepackage{soul} 

\usepackage{mathtools}

\makeatletter
\DeclareRobustCommand\widecheck[1]{{\mathpalette\@widecheck{#1}}}
\def\@widecheck#1#2{%
    \setbox\z@\hbox{\m@th$#1#2$}%
    \setbox\tw@\hbox{\m@th$#1%
       \widehat{%
          \vrule\@width\z@\@height\ht\z@
          \vrule\@height\z@\@width\wd\z@}$}%
    \dp\tw@-\ht\z@
    \@tempdima\ht\z@ \advance\@tempdima2\ht\tw@ \divide\@tempdima\thr@@
    \setbox\tw@\hbox{%
       \raise\@tempdima\hbox{\scalebox{1}[-1]{\lower\@tempdima\box
\tw@}}}%
    {\ooalign{\box\tw@ \cr \box\z@}}}
\makeatother

\usepackage{array}
\newcolumntype{L}[1]{>{\raggedright\let\newline\\\arraybackslash\hspace{0pt}}m{#1}}
\newcolumntype{C}[1]{>{\centering\let\newline\\\arraybackslash\hspace{0pt}}m{#1}}
\newcolumntype{R}[1]{>{\raggedleft\let\newline\\\arraybackslash\hspace{0pt}}m{#1}}

\def\argmin{\mathop{\rm arg\,min}}%

\usepackage{cleveref}

\usepackage{textcomp}

\ifCLASSOPTIONcompsoc
 \usepackage[caption=false,font=normalsize,labelfont=sf,textfont=sf]{subfig}
\else
  \usepackage[caption=false,font=footnotesize]{subfig}
\fi

\DeclareCaptionLabelSeparator{periodspace}{.\quad}

\usepackage[hyphens]{url}

\allowdisplaybreaks 

 \renewcommand{\baselinestretch}{0.97}
\begin{document}
\fontdimen2\font=0.6ex

\title{Real-Time Energy Disaggregation of a Distribution Feeder's Demand Using Online Learning}

\author{Gregory~S.~Ledva,~\IEEEmembership{Student Member, IEEE,}
        Laura~Balzano,~\IEEEmembership{Member, IEEE,}
        and~Johanna~L.~Mathieu,~\IEEEmembership{Member,~IEEE \vspace{-.5cm}}
\thanks{The authors are with the Department of Electrical Engineering \& Computer Science, University of Michigan, Ann Arbor, MI 48109 USA (e-mail: gsledv@umich.edu; girasole@umich.edu; jlmath@umich.edu). This research was funded by NSF Grant \#ECCS-1508943.}
\thanks{IEEE owns the copyright \cite{ledva2017real}}
}

\markboth{\textcopyright 2018 IEEE}%
{Ledva \MakeLowercase{\textit{et al.}}: Real-Time Energy Disaggregation of a Distribution Feeder's Demand Using Online Learning}

\maketitle

\begin{abstract}
Though distribution system operators have been adding more sensors to their networks, they still often lack an accurate real-time picture of the behavior of distributed energy resources such as demand responsive electric loads and residential solar generation. Such information could improve system reliability, economic efficiency, and environmental impact. Rather than installing additional, costly sensing and communication infrastructure to obtain additional real-time information, it may be possible to use existing sensing capabilities and leverage knowledge about the system to reduce the need for new infrastructure. In this paper, we disaggregate a distribution feeder's demand measurements into: 1) the demand of a population of air conditioners, and 2) the demand of the remaining loads connected to the feeder. We use an online learning algorithm, Dynamic Fixed Share (DFS), that uses the real-time distribution feeder measurements as well as models generated from historical building- and device-level data. We develop two implementations of the algorithm and conduct case studies using real demand data from households and commercial buildings to investigate the effectiveness of the algorithm. The case studies demonstrate that DFS can effectively perform online disaggregation and the choice and construction of models included in the algorithm affects its accuracy, which is comparable to that of a set of Kalman filters.

\end{abstract}

\begin{IEEEkeywords}
Online learning, machine learning, energy disaggregation, output feedback, real-time filtering
\end{IEEEkeywords}

\IEEEpeerreviewmaketitle

\section{Introduction}

\IEEEPARstart{D}{istributed} energy resources (DERs) such as demand responsive electric loads and residential solar generation are becoming more common within electricity distribution networks \cite{gtm2015solar, navigant2015direct}. Sensing infrastructure, such as household smart meters, are also becoming more common \cite{lee2015assessment}. However, distribution system operators still often lack an accurate real-time picture of overall DER characteristics such as i) the total power consumption of the air conditioners connected to a distribution feeder, or ii) the total power production of all solar panels installed on a distribution feeder.

Perfect real-time knowledge of DER characteristics requires a sensor at each of the large number (e.g., thousands) of spatially distributed devices and a communication infrastructure capable of reliably transmitting the data at the necessary frequency (e.g., every few seconds). Rather than installing additional, costly metering and communication infrastructure, in this paper, we show that it is possible to estimate real-time DER characteristics using existing sensing capabilities and some knowledge of the underlying system. Specifically, we show how to separate measurements of the net demand served by a distribution feeder into its components in real-time, using knowledge of the physical processes driving load/generation. We refer to this task as {\em feeder-level energy disaggregation}.   


Real-time, feeder-level energy disaggregation can help system operators, utilities, and demand response providers improve power system reliability, economic efficiency, and environmental impact. For example, a system operator can 1) estimate the real-time balancing reserve requirement from its estimate of the production of distributed generation resources; 2) estimate the real-time potential for fault induced delayed voltage recovery (FIDVR) caused by stalling in small motor loads \cite{baone_innovative_2010} from its estimate of the motor load consumption; and 3) optimize conservation voltage recovery (CVR) strategies using its estimate of the mix of constant impedance, constant power, and constant current loads \cite{paul_impact_2013}. A utility can 4) better plan demand response actions by knowing the weather forecast and the real-time portion of weather-dependent loads (e.g., air conditioners, heaters, dehumidifiers). A demand response provider can 5) optimize capacity bids into ancillary services markets using its estimate of the real-time, aggregate, demand-responsive load; and 6) use its estimate of the real-time, aggregate, demand-responsive loads as a feedback signal in load coordination algorithms, e.g., \cite{kara_moving_2013, soudjani_aggregation_2014}. Note that the consumption of demand-responsive loads often needs to be measured for auditing purposes, but the consumption data need not be communicated in real-time.


In this paper, we develop the feeder-level energy disaggregation problem framework and apply an online learning algorithm to separate the active power demand served by a feeder into the active power demand of a population of residential air conditioners and the active power demand of all other loads connected to the feeder. The algorithm \cite{hall2015online} incorporates dynamical system models of arbitrary forms, blending aspects of machine learning and state estimation. Building upon our preliminary work \cite{ledva2015inferring}, the contributions of this paper are to i) frame the feeder-level energy disaggregation problem, ii) adapt the machine learning algorithm in \cite{hall2015online} to the feeder-level energy disaggregation problem, iii) develop a variation of the machine learning algorithm that allows it to include models with different underlying states, iv) demonstrate the performance of the online learning algorithm via a realistic data-driven case study, and v) compare the performance of the algorithm to that of a set of Kalman filters. Beyond \cite{ledva2015inferring}, this paper develops a modified version of the algorithm, compares this modified implementation to a direct implementation of the algorithm, uses only real data (rather than models) to construct the feeder active power signal, uses real data to identify all load models, and compares algorithm performance to that of an aggressive benchmark as opposed to a simple prediction model.



Section~\ref{sec:lit} compares our problem and approach to related problems/work.  Section~\ref{sec:overview} defines the problem framework. Section~\ref{sec:constructionOfPlantSignals} describes the data used to construct the underlying system, and Section~\ref{sec:models} describes the models used within the algorithm. Section~\ref{sec:onlineLearningAlgorithm} summarizes the online learning algorithm and our implementations for the feeder-level energy disaggregation problem. Section~\ref{sec:caseStudies} constructs case studies and summarizes their results. Finally, Section~\ref{sec:conclusions} presents the conclusions. 

\section{Comparison to related problems and work} \label{sec:lit}

The feeder-level energy disaggregation problem combines aspects of building-level energy disaggregation and load forecasting. Building-level energy disaggregation, also referred to as nonintrusive load monitoring\cite{berges2010enhancing}, separates building-level demand measurements into estimates of the demand of  individual or small groups of devices \cite{armel2013disaggregation}. Disaggregation algorithms use data sampled at frequencies ranging from over 1 MHz to 0.3 mHz (i.e., hourly interval data) where higher-frequency data allows separation of more devices \cite{armel2013disaggregation}. The problem is not usually solved online because the goal is long-term energy efficiency decisions such as identification and replacement of faulty appliances and/or load research. Both unsupervised and supervised learning approaches have been proposed, with the latter often using models developed with submetering data.

Load forecasting predicts the total future demand within a given area over time horizons ranging from hours to years \cite{hong2016probabilistic2}. Whereas energy disaggregation typically deals with small load aggregations, load forecasting typically deals with large aggregations, e.g., thousands to millions of loads. For example, forecasting the load served by a distribution transformer is considered a ``small'' forecasting problem \cite{hong2016probabilistic2}.  Very short term load forecasting, corresponding to intraday forecasts, generally uses 15 min to one hour interval data \cite{hong2016probabilistic2,hong2010short}. Smart meter data enables offline development of detailed load models \cite{byers2016how}, which may be used online for operational decisions \cite{taylor2008evaluation}, e.g., for predicting the curtailable load \cite{byers2016how}. However, load forecasting is typically done offline and is typically used for planning.

In contrast to building-level energy disaggregation, feeder-level energy disaggregation involves disaggregating the demand of a large number of loads, e.g., thousands, into a small number of source signals, e.g., two. In contrast to load forecasting, feeder-level energy disaggregation  estimates {\em portions} of the total demand and assumes real-time demand measurements, e.g., taken by SCADA systems at distribution substations, are available on timescales of seconds to minutes. This corresponds to relatively fast sampling for load forecasting and relatively slow sampling for building-level energy disaggregation. In contrast to both building-level energy disaggregation and load forecasting, feeder-level energy disaggregation is done online. However, much like load forecasting and some building-level energy disaggregation approaches, we assume detailed historical load data are available and used offline to construct models. 

Machine learning algorithms have been proposed to address a number of problems in power systems  including security assessment, forecasting, and optimal operation \cite{hatziargyriou2001machine}. A variety of machine learning techniques have been used to forecast load, renewable generation, and prices\cite{negnevitsky2009machine,niu2010power,zhang2005short,sharma2011predicting,teo2015forecasting}. References~\cite{taylor2014index,kalathil2015online,ruelens2016residential,khezeli2016risk,lesage2017learning} apply learning approaches to demand response. However, to the best of our knowledge, this is the first paper to pose and solve the feeder-level energy disaggregation problem, or to apply the approach in \cite{hall2015online} to a power systems problem.

\section{Problem Framework} \label{sec:overview}

\begin{figure} 
\centering 
\includegraphics[scale=1]{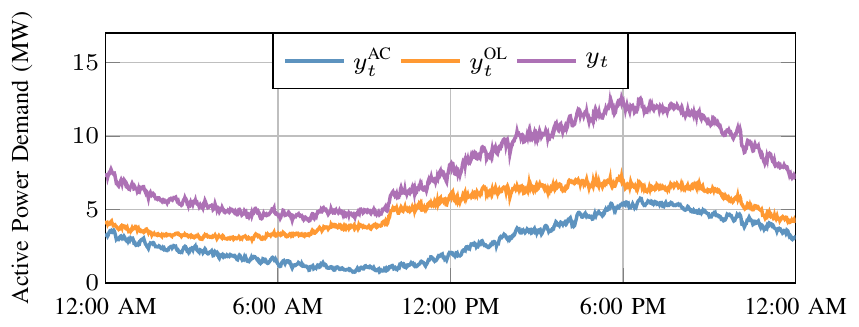}
\vspace{-8pt}		
\caption{Example time series of $y_t$ and its components $y^\text{OL}_t$ and $y^\text{AC}_t$.} 	
\vspace{-5pt}		
\label{fig:problemIntro}
\end{figure}

We assume that a power system entity (e.g., a system operator, utility, or third-party company) has access to real-time measurements of the electricity demand served by a distribution feeder. The power system entity is interested in separating these measurements into two components in real-time, i.e., at each time-step. The first component is the power demand of a population of residential air conditioners served by the feeder, referred to as the ``AC demand.'' Air conditioners generally draw power periodically to maintain a building's indoor temperature within a range centered at a user-defined temperature set-point. The AC demand varies in time due to each air conditioner's power cycling, weather-related influences, and building occupant influences. The second component is the power demand of the other loads on the feeder, referred to as the ``OL demand,'' which we assume includes both residential and commercial loads. Figure~\ref{fig:problemIntro} displays example time series for the measured total demand $y_t$, the AC demand $y_t^\text{AC}$, and the OL demand $y_t^\text{OL}$ over a day. We measure $y_t$ at each time-step and try to estimate $y_t^\text{AC}$ and  $y_t^\text{OL}$ at each time-step as each measurement arrives. 

\begin{figure}
\centering
\subfloat[Real-time estimation mode]{
\includegraphics[scale=1]{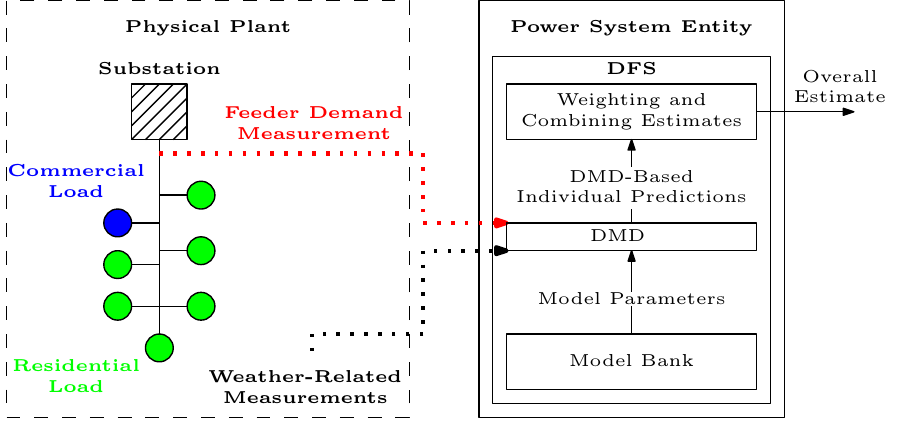}
\label{fig:systemOverviewA}
}
\\
\subfloat[Offline model generation mode]{
\includegraphics[scale=1]{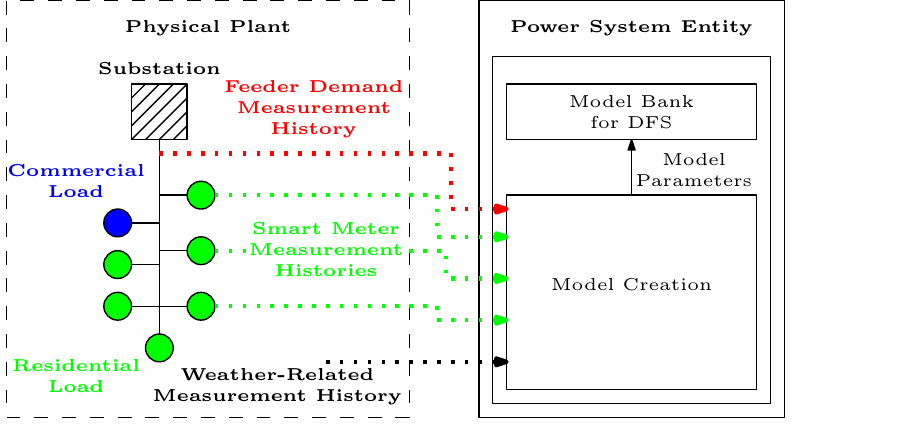}
\label{fig:systemOverviewB}
}
\caption{Problem framework: real-time and offline modes.} 
\label{fig:systemOverview}
\vspace{-15pt}
\end{figure}

The power system entity has two distinct modes of operation. The first is the real-time estimation mode depicted in Fig.~\ref{fig:systemOverviewA}. The second is the offline model generation mode, depicted in Fig.~\ref{fig:systemOverviewB}.  During real-time operation, we assume that the power system entity has access to active power measurements corresponding to the demand served by the distribution feeder as well as weather-related measurements. The power measurements are time-averaged active power demands over one min intervals, and they are the sum of the AC and OL demand. The weather-related measurements could include, for example, temperature and humidity, and can be obtained from existing weather sensors; load-specific weather monitoring is not required.

Model generation occurs offline using historical smart meter, feeder, and weather data. To apply DFS to feeder-level energy disaggregation we assume real-time measurements of the demand components are unavailable, but models of the components are available. These models could be created using a variety of techniques (e.g., via system identification using historical measurements obtained from the same system or a different system, or using analytical methods and parameters from the literature). In this work, we assume that smart meters are installed at all houses, and they enable the collection of household-level demand measurements at one minute intervals. The smart meters' communication limitations \cite{armel2013disaggregation} make real-time communication of this information infeasible, and so we assume that it is only available offline for prior days. Because we do not need real-time, device-level demand measurements, we assume that historical device-level demand estimates can be obtained offline from the historical household-level measurements either by applying non-intrusive load monitoring (NILM) algorithms or by using information from communicating or advanced thermostats, which are becoming more common within residences. These thermostats can measure and record the on/off mode of a residence's AC unit and measurement histories can be used to estimate the power draw of these devices. The resulting device-level demand estimates may not be exact, but they are accurate enough to be used within the computation of model parameters. As a result, we use historical, device-level measurements to construct the AC demand models. We also assume that the power system entity has access to historical feeder and weather data. Once the models are formed, they are used along with the real-time measurements to estimate the AC and OL demand. 

An online learning algorithm, Dynamic Mirror Descent (DMD) \cite{hall2015online}, uses a single model to generate predictions of the total demand, a loss function to penalize errors between the predicted and measured total demand, and a convex optimization formulation to adjust this prediction based on the measured total demand. Dynamic Fixed Share (DFS) \cite{hall2015online}, uses DMD within the Fixed Share Algorithm \cite{herbster1998tracking} to include predictions from a bank of models. Specifically, DFS applies DMD separately to each model and uses a weighting algorithm to associate a weight with each model's adjusted prediction before combining the predictions into an overall estimate. In DFS, these models are weighted based on their prediction accuracy -- better prediction-measurement matching leads to larger weighting and more influence in the overall prediction. 

Rather than predicting the AC demand using a single load forecast, the proposed approach has two main advantages. First, in DMD, the AC and OL demand predictions are adjusted in real-time based on the real-time, realized feeder demand. This feedback improves future predictions; in contrast, load forecasting is open-loop. It is necessary to predict both the AC and OL demand since only the total demand is measured. If only the AC demand is predicted, the prediction cannot be adjusted in real-time because measurements of the realized AC demand are not available in real-time. Second, the DFS algorithm can incorporate a number of AC demand predictions into an overall AC demand prediction. Predictions associated with prediction methods that have performed well recently are weighted more heavily and the weights evolve over time so different predictors will be preferred at different times. The algorithm is described in detail in Section~\ref{sec:onlineLearningAlgorithm}, but first we describe the construction of the underlying physical system, i.e., the plant, used within the case studies and the models used within the algorithm. 

\section{Construction of Plant} 
\label{sec:constructionOfPlantSignals}

In this section, we detail the methods used to form the AC and OL demand time series and the associated weather time series over one day. 
These time series incorporate data from real households, the devices within those households, commercial buildings, and nearby weather stations. The data for individual, residential air conditioners are summed to form the AC demand, the data for household non-AC devices are summed to form the residential OL demand, and the data for commercial building demand signals are summed and scaled to form the commercial, OL demand signal. Lastly, the outdoor temperature data consists of real data from nearby weather stations, and the data is interpolated to make it applicable on the time-steps used within the problem scenario. These time series are then used as the plant, i.e., the underlying physical system or the ground-truth signals. The time series for a day consist of $n^\text{steps}$ one-minute time-steps with $ t=0 $ at 12:00 AM. 
Because we were unable to find sufficient data from a single location/day, we use demand and weather data from a variety of sources.

We use feeder model R5-25.00-1 from GridLAB-D's feeder taxonomy \cite{schneider2008modern} to set the average residential and commercial demand on the feeder to $5.8$ MW and $2.1$ MW, respectively. Ignoring network losses (which, if included, would be treated as part of the OL demand), the total feeder demand measurements are the sum of the AC and OL demand, i.e., $y_t = y_t^\text{AC} + y_t^\text{OL}$, where $ y_t^\text{OL}$ is the sum of the other residential demand and the commercial demand $y^\text{OL}_t = y^\text{OL,res}_t + y^\text{OL,com}_t$. 

Both $y^\text{AC}_t$ and $y^\text{OL,res}_t$ are constructed using residential data from the Pecan Street Dataport \cite{pecanstreet}. The data consists of historical one min interval household- and device-level demand  measurements for a set of single family homes in Texas. Daily household demand signals were randomly drawn with replacement and added together until the total residential signal's mean matched that of the feeder model, resulting in $2,499$ total houses. To construct the AC demand signal, we summed the demand of each household's primary air conditioner and air blower unit. Note that some houses have no/multiple air conditioner and air blower units. We assume that only one unit per household contributes to the AC demand, resulting in $2,269$ units. The remaining demand is the residential OL demand. 

The commercial data consists of 4 second interval whole-building demand measurements from two buildings in California, a municipal building and a big box retail store. We summed the demand of the two buildings, and then scaled the sum by 2.61 to match the average commercial demand of the feeder model. We also down-sampled the data to one min intervals by averaging the values over each minute. 

The plant's weather data is constructed from data obtained from the Pecan Street Dataport \cite{pecanstreet} and the National Climatic Data Center \cite{noaa2009nndc}. The Pecan Street weather data corresponds to the residential demand. It consists of the outdoor air temperature for Austin, TX, and it is sampled at one hour intervals. We linearly interpolate the data down to one min intervals.  The NOAA weather data corresponds to the commercial demand. It consists of outdoor temperature data from the Concord, CA weather station, sampled at one hour intervals. Again, we linearly interpolate the data down to one min intervals. All weather data was taken from the same day as the demand data.


\section{System Models}
\label{sec:models}

In this section, we describe the models used to generate predictions of the AC and OL demands. These models are generated offline, using historical data, and then used within the online learning algorithm detailed in Section~\ref{sec:onlineLearningAlgorithm}. 
The historical demand signals were constructed in the same manner as described in Section~\ref{sec:constructionOfPlantSignals}, using the same combination of houses as used to construct the plant signals. 
The OL demand is modeled using two different linear regression methods as detailed in Section~\ref{sec:uncontrollableModels}, and the AC demand is modeled using several linear dynamic systems as well as a linear regression method as detailed in Section~\ref{sec:controllableModels}. Note that other models may prove to be more accurate on average than the models that are used within this work. However, the intended objective within this work is to use an array of models (including some that are known to be overly simple or less accurate) to investigate the performance of DFS on the feeder-level energy disaggregation problem.

\subsection{OL Demand Models}
\label{sec:uncontrollableModels}
\begin{figure} 
\centering 
\includegraphics[scale=1]{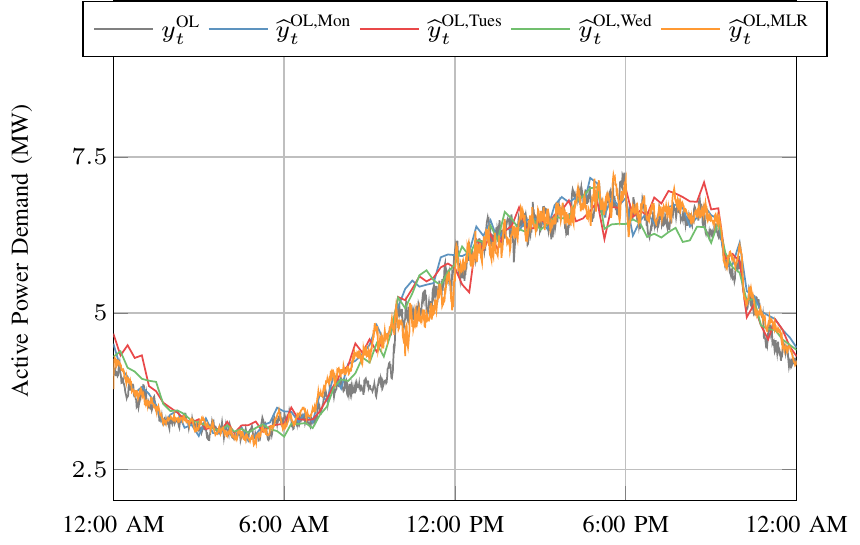}
\vspace{-5pt}		      
\caption{Example OL demand and several OL demand model predictions.} 	
\vspace{-10pt}		
\label{fig:uncontrollableModels}
\end{figure}

We use two types of regression models to predict the OL demand: time-of-day (TOD) regression models and a multiple linear regression (MLR) model. Figure~\ref{fig:uncontrollableModels} displays  $y^\text{OL}_t$ for a simulated day, several TOD regression model predictions, e.g., $\widehat{y}^\text{OL,Mon}_t$, 
and the MLR model prediction $\widehat{y}^\text{OL,MLR}_t$. We next describe the construction of these models. 

\subsubsection{TOD Regression Models}
The TOD regression model is a lookup table where an OL demand prediction is generated for each minute of the day based on the OL demand of a single day in the past
\begin{align}
\widehat{y}^\text{OL,TOD}_t =  \alpha^\text{OL,TOD}_k  =  \alpha^\text{OL,TOD}\;  x^\text{OL,TOD}_t.
\end{align}
Whereas $t$ indexes overall time-steps, $k$ indexes the time of day in minutes, i.e., $k = 0$ for 12:00 AM and $k = 60$ for 1:00 AM. The scalar $\alpha^\text{OL}_k$ corresponds to the predicted OL demand value for time-of-day $k$, $\alpha^\text{OL,TOD}$ is a row vector containing all $\alpha^\text{OL}_k$ values, and $x^\text{OL,TOD}_t$ is a column vector that selects the appropriate $\alpha^\text{OL,TOD}$ based on the corresponding time of day for $t$. We generate $\alpha^\text{OL,TOD}$ by smoothing the OL demand signal of a previous day using a piecewise linear and continuous, least-squares fit. Each linear segment corresponds to a 15 min interval of the historical data. We generate one TOD regression model for each weekday, and the models are denoted $\Phi^\text{OL,Mon}$, $\Phi^\text{OL,Tues}$, $\Phi^\text{OL,Wed}$, $\Phi^\text{OL,Thurs}$, and $\Phi^\text{OL,Fri}$. Their corresponding predictions are $\widehat{y}^\text{OL,Mon}_t$, $\widehat{y}^\text{OL,Tues}_t$, $\widehat{y}^\text{OL,Wed}_t$, $\widehat{y}^\text{OL,Thurs}_t$, and $\widehat{y}^\text{OL,Fri}_t$, respectively. 


\subsubsection{MLR Model}
\label{sec:uncontrollableMLR}
The MLR model of the OL demand is denoted $\Phi^\text{OL,MLR}$, and it uses input features that include calendar-based variables, e.g., the day of the week, as well as weather-based variables, e.g., the outdoor temperature, to generate an OL demand prediction. We split the MLR model into two distinct components: one model for the commercial demand and one model for the residential OL demand since the underlying data corresponds to different geographic areas and time periods. The overall MLR model of the OL demand is then the sum of the predicted residential OL demand $\widehat{y}^\text{OL,res}_t$ and the predicted commercial demand $\widehat{y}^\text{OL,com}_t$, i.e., 
\begin{align}
\widehat{y}^\text{OL,MLR}_t = & \widehat{y}^\text{OL,res}_t \; + \; \widehat{y}^\text{OL,com}_t \\
=&  \beta^\text{OL,res} \;  x^\text{OL,res}_t \; + \; \gamma^\text{OL,com} \;  x^\text{OL,com}_t,
\end{align}
where the row vectors $\beta^\text{OL,res}$ and $\gamma^\text{OL,com}$ are regression parameters for the residential OL demand and the commercial demand, respectively. The column vectors $x^\text{OL,res}_t$ and $x^\text{OL,com}_t$ are the corresponding input features. 

The MLR model for the residential OL demand uses input features $x^\text{OL,res}_t = \begin{bmatrix} 
(x_t^\text{TOW})^\text{T} & T^\text{TX}_t & y_{t-1}
\end{bmatrix}^\text{T}$ where $x_t^\text{TOW}$ is an indicator vector for the time of week in minutes, $T^\text{TX}_t$ is the outdoor temperature for Austin, TX, and $y_{t-1}$ is the measured total demand of the previous time-step. 
The commercial regression model corresponds to ``Baseline Method 1'' from \cite{mathieu2010characterizing};  
it uses input features  $x^\text{OL,com}_t = \begin{bmatrix}
 (x^\text{TOW}_t)^\text{T} & \; T^\text{CA}_t \cdot(x^\text{TOW}_t)^\text{T} \end{bmatrix}^\text{T}$ where $T^\text{CA}_t$ is the outdoor temperature for Concord, CA and $T^\text{CA}_t \cdot(x^\text{TOW}_t)$ is a vector that associates the temperature to the corresponding time of week. 


\subsection{AC Demand Models}
\label{sec:controllableModels}
\begin{figure}
\centering
\includegraphics[scale=1]{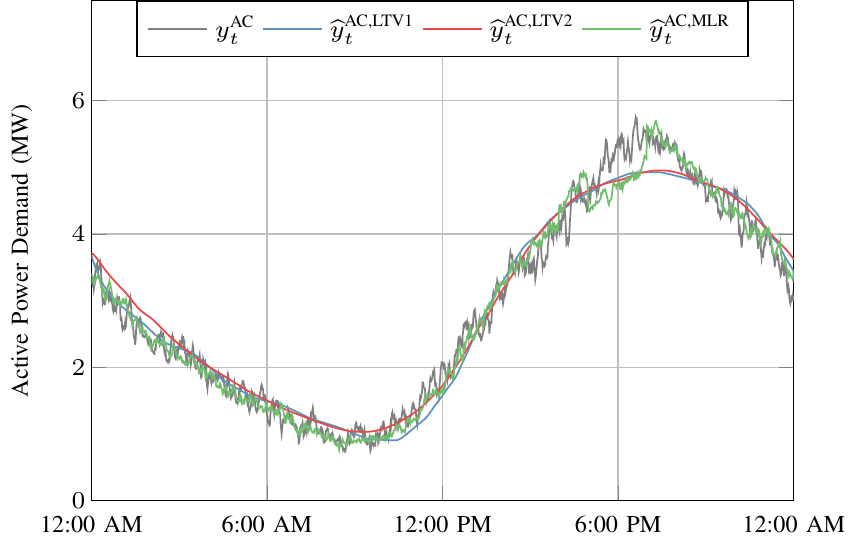}
\vspace{-5pt}        
\caption{Examples AC demand and several AC demand model predictions.} 	
\label{fig:controllableModels}
\vspace{-10pt}        
\end{figure}

We use three types of models to predict the AC demand: a MLR model, linear time invariant (LTI) system models, and linear time varying (LTV) system models. Figure~\ref{fig:controllableModels} 
displays  $y^\text{AC}_t$ for a simulated day, several LTV model predictions, e.g., $\widehat{y}^\text{AC,LTV1}_t$, 
and the MLR regression model prediction $\widehat{y}^\text{AC,MLR}_t$. We next describe the construction of these models.

\subsubsection{MLR Model}
\label{sec:mlrAcModel}
The MLR model of the AC demand, denoted $\Phi^\text{AC,MLR}$, is similar to the MLR model in Section~\ref{sec:uncontrollableMLR} with different input features $x^\text{AC,MLR}_t = \begin{bmatrix} (x_t^\text{TOW})^\text{T} & T^\text{TX}_{t-\tau^\text{l}} & (T^\text{TX}_{t-\tau^\text{l}})^2 & (T^\text{TX}_{t-\tau^\text{l}})^3 & (T^\text{TX}_{t-\tau^\text{l}})^4
 \end{bmatrix}^\text{T}$, where $T^\text{TX}_{t-\tau^\text{l}}$ is the temperature in Austin, TX from $\tau^\text{l}$ time-steps ago and $\tau^\text{l}$ is the time lag that maximizes the cross correlation between the historical AC demand signal and temperature signal (119 min for our plant). 

\subsubsection{LTI Models}
\label{sec:ltiModels}
We construct a set of LTI models $\mathcal{M}^\text{LTI}$, originally developed in \cite{mathieu2013state,kalsi2012development}. As in \cite{mathieu2015arbitraging}, each model within the set captures the aggregate behavior of the population of air conditioners at outdoor temperature $T^m$ and has the form
\begin{align}
\widehat{x}^{\text{LTI},m}_{t+1} =& \:  A^{\text{LTI},m} \; \widehat{x}^{\text{LTI},m}_t \label{eq:ltiStateUpdate}\\
\widehat{y}^{\text{AC,LTI,}m}_t =& \: C^{\text{LTI},m} \; \widehat{x}^{\text{LTI},m}_t ,
\end{align} 
with $m \in \mathcal{M}^\text{LTI} = \{ 1, \dots, N^\text{LTI} \}$. The state vector $\widehat{x}_t^{\text{LTI},m} \in \mathbb{R}^{N^\text{x} \times 1} $ consists of the portion of the air conditioners within each of $N^\text{x}$ discrete states. In this paper, we use one state to represent the portion of air conditioners that are drawing power and another to represent those that are not, i.e., $N^\text{x} = 2$. The state transition matrix, $A^{\text{LTI},m} \in \mathbb{R}^{N^\text{x} \times N^\text{x}}$, is a transposed Markov transition matrix. Its entries capture the probabilities that air conditioners maintain their current state or transition to the other state during the time-step. The output matrix $C^{\text{LTI},m}$ estimates the AC demand $\widehat{y}^{\text{AC,LTI,}m}_t$ from the portion of air conditioners that are drawing power, i.e., $C^{\text{LTI},m} = N^\text{ac} \overline{P}^m \begin{bmatrix} 0 & 1\end{bmatrix}$, where $\overline{P}^m$ is a parameter approximating of the average power draw of air conditioners drawing power and $N^\text{ac}$ is the number of air conditioners, which we assume is known. 

To identify $A^{\text{LTI},m}$ and $C^{\text{LTI},m}$ for all $m$, we first define a set of $N^\text{LTI}$ evenly spaced temperatures $\mathcal{T}^\text{temps} = \left\{T^\text{min}, \dots , T^\text{max}\right\}$ and denote the $m$-th temperature of the set as $T^m$.  The difference between successive temperatures $T^m$ and $T^{m+1}$ is $\Delta T$. Matrices $A^{\text{LTI},m}$ and $C^{\text{LTI},m}$ are constructed using power demand signals from each air conditioner corresponding to periods when $T^m-\frac{\Delta T}{2}  \; \leq \;  T^\text{TX}_{t-\tau^\text{l}} \; < \; T^m+\frac{\Delta T}{2}$. Some heuristics were used to exclude anomalous high or low power demand measurements. Parameter $\overline{P}^m$ is set as the average power draw of air conditioners that are drawing power. The four entries of $A^{\text{LTI},m}$ were determined by checking whether an air conditioner 1) started drawing power, 2) stopped drawing power, 3) continued to draw power, or 4) continued to not draw power during each time-step. The occurrences for each case were counted for every air conditioner at every time-step and the totals were placed into their respective entries in $A^{\text{LTI},m}$, and then each column was normalized so that the sum of the column entries was 1. In our case studies, we construct an LTI model for each integer temperature in the set  $\left\{74, \dots,99 \right\}$ $^{\circ}\text{F}$. If the outdoor temperature lies outside of this range, we use the model corresponding to the closest temperature. 

\subsubsection{LTV Models}
\label{sec:delayedTempLtv}
We use two LTV models. The first $\Phi^\text{AC,LTV1}$ uses the delayed temperature and 
has the form 
\begin{align}
\widehat{x}^\text{LTV1}_{t+1} =& \:  A^\text{LTV1}_t \; \widehat{x}^\text{LTV1}_t \label{eq:ltv1StateUpdate}\\
\widehat{y}_t^\text{AC,LTV1} =& \: C^\text{LTV1}_t \; \widehat{x}^\text{LTV1}_t \label{eq:ltv1Output},
\end{align}
where $A^\text{LTV1}_t$ and $C^\text{LTV1}_t$ are generated by linearly interpolating the matrix entries based on $T^\text{TX}_{t-\tau^\text{l}}$. 
The second $\Phi^\text{AC,LTV2}$ uses a moving average of the past temperature over $\tau^\text{w}$ time-steps to generate the prediction $\widehat{y}_t^\text{AC,LTV2}$.
We chose $\tau^\text{w}$ to be the value that maximizes the cross correlation between the historical moving average temperature and the historical AC demand signal (270 min for our plant). 
When evaluating either LTV model, if the temperature lies outside of the range used to generate the model, we extrapolate using the difference between the nearest two models. 

\section{Online Learning Algorithm} \label{sec:onlineLearningAlgorithm}
In this section, we first summarize the DFS algorithm developed in \cite{hall2015online} and then describe two algorithm implementations, one inspired by DFS and one a direct implementation of it. DFS incorporates DMD, also developed in \cite{hall2015online}, into the Fixed Share algorithm originally developed in \cite{herbster1998tracking}. The Fixed Share algorithm combines a set of predictions that are generated by independent experts, e.g., models,  into an estimate of the system parameter using the experts' historical accuracy with respect to observations of the system. DMD extends the traditional online learning framework by incorporating dynamic models, enabling the estimation of time-varying system parameters (or states). DFS uses DMD, applied independently to each of the models, as the experts within the Fixed Share algorithm.

\subsection{The DFS Algorithm}
\label{sec:DfsAlgorithm}

The objective of DFS is to form an estimate $\widehat{\theta}_t \in \Theta$ of the dynamic system parameter $\theta_t \in \Theta$ at each discrete time-step $t$ where $\Theta \subset \mathbb{R}^p$ is a bounded, closed, convex feasible set. The underlying system produces observations, i.e., measurements, $y_t \in \mathcal{Y}$ at each time-step after the prediction has been formed, where $\mathcal{Y} \subset \mathbb{R}^q$ is the domain of the measurements. From a control systems perspective, this is equivalent to a state estimation problem where $\theta_t$ is the system state. 

DFS uses a set of $N^\text{mdl}$ models defined as $\mathcal{M}^\text{mdl} = \{ 1, \, \dots \,, N^\text{mdl} \}$ to generate the estimate $\widehat{\theta}_t$. To do this, DFS applies the DMD algorithm to each model, forming predictions $\widehat{\theta}^m_{t}$ for each $m \in \mathcal{M}^\text{mdl}$. DMD is executed in two steps (similar to a discrete-time Kalman filter): 1) an observation-based update incorporates the new measurement into the parameter prediction, and 2) a model-based update advances the parameter prediction to the next time-step. DFS then uses the Fixed Share algorithm to form the estimate $\widehat{\theta}_{t}$ as a weighted combination of the individual model's DMD-based predictions. A weighting algorithm computes the weights based on each model's historical accuracy with respect to the observations $y_t$. Models that perform poorly have less influence on the overall estimate.
The DFS algorithm is \cite{hall2015online}
\begin{align}
\widetilde{\theta}^m_t = & \;   \underset{\theta \in \Theta}{\argmin} \; \eta^\text{s} \left\langle \nabla \ell_t(\widehat{\theta}^m_t), \; \theta \right\rangle + D\left( \theta \| \widehat{\theta}^m_t \right) \label{eq:gradientUpdateDFS} \\
\widehat{\theta}^m_{t+1}  = & \Phi^m (\widetilde{\theta}^m_t) \label{eq:modelUpdateDFS} \\
w^m_{t+1} = & \frac{\lambda}{N^\text{mdl}} + (1- \lambda)  \frac{w^m_t \, \exp\left(-\eta^r  \, \ell_t \left(\widehat{\theta}^m_t \right)\right)}{\sum_{j=1}^{N^\text{mdl}} w^j_t \, \exp\left(-\eta^r  \, \ell_t \left(\widehat{\theta}^j_t \right)  \right)} \label{eq:wHat}
\end{align}
for each $m \in \mathcal{M}^\text{mdl}$, and
\begin{equation}
\widehat{\theta}_{t+1} =  \sum_{m \in \mathcal{M}^\text{mdl}} w^m_{t+1} \; \widehat{\theta}_{t+1}^m ,\label{eq:thetaDFS}
\end{equation}
where each term is defined below. DMD is applied to each model in \eqref{eq:gradientUpdateDFS} and \eqref{eq:modelUpdateDFS} to form the expert predictions, where \eqref{eq:gradientUpdateDFS} is a convex program that constructs the measurement-based update to the previous prediction and \eqref{eq:modelUpdateDFS} is the model-based advancement of the adjusted prediction. The Fixed Share algorithm consists of \eqref{eq:wHat} and \eqref{eq:thetaDFS}, where \eqref{eq:wHat} computes the weights and \eqref{eq:thetaDFS} computes the estimate as a weighted combination of the individual experts' estimates. We note that the Fixed Share algorithm's updates are independent of the dynamics and only use the experts' predictions and their resulting losses. 

In \eqref{eq:gradientUpdateDFS}, we minimize over the variable $\theta$, $\eta^\text{s} > 0$ is a step-size parameter, and $\langle \cdot, \cdot \rangle$ is the standard dot product. The value $ \nabla \ell_t(\widehat{\theta}_t)$ is a subgradient of the convex loss function $\ell_t : \Theta \rightarrow \mathbb{R}$, which penalizes the error between the predicted and observed values $y_t$ using a known, possibly time-varying, function $h_t : \Theta \rightarrow \mathcal{Y}$ that maps $\theta_t$ to an observation, i.e., $y_t = h_t(\theta_t )$, to form predictions of the measurements. An example loss function is $\ell_t(\widehat{\theta}_t) = \lVert C \widehat{\theta}_t - y_t \rVert_2^2$ where the matrix $C$ is $h_t(\cdot)$. In \eqref{eq:modelUpdateDFS}, the function $\Phi^m(\cdot)$ applies model $m$ to advance the adjusted estimate $\widetilde{\theta}^m_t$ in time. Each $\Phi^m(\cdot)$ can have arbitrary form and time-varying parameters. In \eqref{eq:wHat}, the weight associated with model $m$ at time-step $t$ is $w^m_{t}$, $\lambda \in (0,1)$ determines the amount of weight that is shared amongst models, and $\eta^r $ influences switching speed. The weight for model $m$ is based on the loss of each model and the total loss of all models.
The term $\eta^\text{s} \langle \nabla \ell_t(\widehat{\theta}_t), \; \theta \rangle$ captures the alignment of the variable $\theta$ with the positive gradient of $\ell_t(\widehat{\theta_t})$. To minimize this term alone, we would choose $\theta$ to be exactly aligned with the negative gradient direction. The term $D (\theta \| \widehat{\theta}_t )$ is a Bregman divergence that penalizes the deviation between the new variable $\theta$ and the old variable $\widehat{\theta}_t$. 
For simplicity, we have excluded regularization within \eqref{eq:gradientUpdateDFS}, which DMD readily incorporates \cite{hall2015online}.

\subsection{Algorithm Implementations}
\label{sec:dmdImplementation}

We next describe two algorithm implementations to update the expert predictions. First, we describe an implementation that uses the concept of DMD 
but it is not a direct implementation of DMD. This method treats the models as black boxes and adjusts only their output, i.e., the OL and AC demand predictions, using the measured and predicted total feeder demand. Second, we describe a direct implementation of DMD, which updates the state $x_t$ of the LTI and LTV AC demand models. 
In the following, the total demand model is $\Phi(\cdot) = \{\Phi^\text{AC}(\cdot), \Phi^\text{OL}(\cdot) \}$ where $\Phi^\text{AC}(\cdot)$ is an AC demand model and $\Phi^\text{OL}(\cdot)$ is an OL demand model, with predictions $\widehat{y}_t^\text{AC}$ and $\widehat{y}_t^\text{OL}$, respectively. 

\subsubsection{Update Method 1}
\label{sec:dmd1}
The models used within this paper have different underlying parameters, dynamic variables, and/or structures, which makes it difficult to define a common $\theta_t$ across all of the models used. Therefore, we develop a variation of the DMD algorithm that adjusts the demand predictions directly, rather than applying the updates to quantities influencing the demand predictions. This allows us to include a diverse set of models. Specifically, we modify the DMD formulation to
\begin{align}
\widehat{\kappa}_{t+1} = & \;   \underset{\theta \in \Theta}{\argmin} \; \eta^\text{s} \left\langle \nabla \ell_t(\widehat{\theta}_t), \; \theta \right\rangle + D\left( \theta \| \widehat{\kappa}_{t} \right)  
\label{eq:method1Kappa} \\
\widecheck{\theta}_{t+1}  = & \Phi (\widecheck{\theta}_{t}) \label{eq:method1check}\\
\widehat{\theta}_{t+1} = & \widecheck{\theta}_{t+1} + \widehat{\kappa}_{t+1}. \label{eq:method1Combine}
\end{align}
The AC and OL demand models generate their predictions independently from one another, and so \eqref{eq:method1Combine} can be rewritten as
\begin{align}
\widehat{\theta}_{t+1} = \Phi(\widecheck{\theta}_t) + \widehat{\kappa}_{t+1} 
=  \begin{bmatrix}
\Phi^\text{AC}(\widecheck{\theta}_t) \\
\Phi^\text{OL}(\widecheck{\theta}_t)
\end{bmatrix} + \widehat{\kappa}_{t+1}.
\end{align} 
The convex program \eqref{eq:method1Kappa} is now used to update a value $\widehat{\kappa}_{t}$ that accumulates the deviation between the predicted and actual measurements. The model-based update \eqref{eq:method1check} computes an open-loop prediction $\widecheck{\theta}_{t+1}$, meaning that the measurements do not influence $\widecheck{\theta}_{t+1}$. The measurement-based updates and model-based, open-loop predictions are combined in \eqref{eq:method1Combine}. In contrast, DMD uses a closed-loop model-based update where the convex program adjusts the parameter estimate to $\widetilde{\theta}_t$, which is used to compute the next parameter estimate $\widehat{\theta}_{t+1}$.

In this method, we define $\theta_t$ as the AC and OL demand, i.e., $\theta_t = \begin{bmatrix}
y^\text{AC}_t & y^\text{OL}_t \end{bmatrix}^\text{T}$. The mapping from the parameter to the measurement is $h_t(\theta_t) = C_t \theta_t$ where the matrix $C_t = \begin{bmatrix} 1 & 1 \end{bmatrix}$. While the mapping and matrix are time-invariant, they may be time-varying in Section~\ref{sec:dmd2}, and so we use the more general notation. We choose the loss function as 
$\ell_t(\widehat{\theta}_t) = \frac{1}{2} \lVert C_t \widehat{\theta}_t - y_t \rVert_2^2$ and the divergence as $D(\theta \rVert \widehat{\kappa}_{t} ) = \frac{1}{2} \lVert  \theta - \widehat{\kappa}_{t} \rVert_2^2$. We can then write \eqref{eq:method1Kappa} in closed form as 
\begin{align}
\widehat{\kappa}_{t+1} &= \widehat{\kappa}_{t} +  \eta^\text{s} C_t^T \left(y_t - C_t \widehat{\theta}_t \right). \label{eq:imp1}
\end{align}

\subsubsection{Update Method 2}
\label{sec:dmd2}
This method applies only to dynamic system models with dynamic states, i.e., in this paper the LTI or LTV AC demand models, which have dynamic states $x_t$. We set $\theta_t = \begin{bmatrix}
x_t^\text{T} & y_t^\text{OL} \end{bmatrix}^\text{T}$, where $x_t$ is $\widehat{x}_t^{\text{LTI},m}$ in \eqref{eq:ltiStateUpdate}, $\widehat{x}_t^{\text{LTV1}}$ in \eqref{eq:ltv1StateUpdate}, or $\widehat{x}_t^{\text{LTV2}}$ in an update equation similar to \eqref{eq:ltv1StateUpdate}. The mapping from the parameter to the measurement is then $C_t = \begin{bmatrix} C_t^\text{AC} & 1 \end{bmatrix}$ where $C_t^\text{AC}$ is the output matrix of the LTI or LTV AC demand model, i.e., $C^{\text{LTI},m}$, $C^\text{LTV1}_t$, or $C^\text{LTV2}_t$. Defining the system parameter in this way allows us to update the dynamic states of the LTI and LTV AC demand models, rather than just the output as in Update Method 1.
The model-based update is then 
\begin{align}
\widehat{\theta}_{t+1} = & \begin{bmatrix}
\Phi^\text{AC}(\widetilde{\theta}_t) \\
\Phi^\text{OL}(\widecheck{\theta}_t)
\end{bmatrix} + 
\begin{bmatrix}
0 & 0 \\
0 & 1
\end{bmatrix}
\widehat{\kappa}_{t+1},
\end{align}
where we update the AC demand model using the adjusted parameter estimate, as in DMD. Because the OL demand models do not include dynamic states, we continue to update their estimates according to Update Method 1. 
We again use \eqref{eq:imp1} as the measurement-based update. 


\section{Case Studies}
\label{sec:caseStudies}
In this section, we define the scenarios, describe the benchmark, summarize the parameter settings, and present the results.  In \cite{hall2015online}, performance bounds for DMD and DFS were established in terms of a quantity called regret. Regret is the total (or cumulative) loss of an online learning algorithm's prediction sequence versus that of a comparator sequence, often a best-in-hindsight offline algorithm. In \cite{hall2015online}, the DMD regret bound uses a comparator that can take on an arbitrary sequence of values from the feasible domain $\Theta$. The DFS regret bound uses a comparator that chooses the best-in-hindsight possible sequence of models chosen from the same model collection used by DFS, where the number of model switches is a predefined number. In lieu of developing formal performance bounds for the given problem scenario, we benchmark the algorithms' performance using Kalman filters, which is described in Section~\ref{sec:benchmark}. Future work will investigate regret bounds for our particular problem.

\subsection{Scenario Definitions}


We define the three sets of models for use within DFS: 
\begin{enumerate}
\item $\mathcal{M}^\text{Full}$, all of the models developed in Section~\ref{sec:models}, i.e., every combination of AC and OL demand models from the AC demand model set $\mathcal{M}^\text{AC,Full} = \{ \mathcal{M}^\text{LTI}, \: \Phi^\text{AC,MLR}, \: \Phi^\text{AC,LTV1}, \: \Phi^\text{AC,LTV2} \} $ and the OL demand model set $\mathcal{M}^\text{OL} = \{ \Phi^\text{OL,Mon}, \: \Phi^\text{OL,Tues}, \: \Phi^\text{OL,Wed}, \: \Phi^\text{OL,Thurs}, \: \Phi^\text{OL,Fri},\: \Phi^\text{OL,MLR} \} $;
\item $\mathcal{M}^\text{Red}$, a reduced set that excludes the LTI models, which are not accurate over the course of the day; 
\item $\mathcal{M}^\text{KF}$, a further reduced set that excludes the MLR AC demand model, which can not be used in a Kalman filter; 
\end{enumerate}
Since the Update Method 2 is only applicable to the LTI and LTV AC demand models, case studies using Update Method 2 apply the method to all applicable model combinations and otherwise use Update Method 1. 

\subsection{Kalman filter benchmark}
\label{sec:benchmark}
A set of Kalman filters are used to establish a  benchmark for the DFS algorithm.  A Kalman filter uses measurements, an assumed system model, and known statistics of random variables, which are assumed to be zero-mean and normally distributed, to estimate the value of dynamic system parameters, i.e., the system state, at each time-step. Additional background on Kalman filters can be found in \cite{grewal2015kalman}.

We use the LTV AC demand models within the Kalman filters. For each LTV model, the covariance of the process noise is computed using a week of historical data, where the true state is generated using the measured AC demand and the LTV model's matrices. The Kalman filter estimates the state of the AC demand model, i.e., $\theta_t = x_t$ where $x_t$ is $\widehat{x}_t^{\text{LTV1}}$ or $\widehat{x}_t^{\text{LTV2}}$, using output pseudo-measurements of the AC demand $\widetilde{y}^\text{AC}_t = y_t-\widehat{y}_t^\text{OL}$. We assume that $y_t$ is noise-free, but $\widetilde{y}^\text{AC}_t$ is noisy due to OL demand prediction error. The covariance of the measurement errors depends on the OL demand model used, and is computed for each model using a week of historical errors. 

We run one Kalman filter for each model pair in the set $\mathcal{M}^\text{KF}$. We compare the performance of the DFS algorithm to that of the best Kalman filter (BKF), which takes the lowest \emph{ex post} root mean squared error (RMSE) achieved by a Kalman filter within the set $\mathcal{M}^\text{KF}$, and the average Kalman filter (AKF), which is the average RMSE  across all of the Kalman filters. 



\subsection{Data}
We test the methods on data from Aug 3-5, 10-14, 17, and 18, where the commercial data is from 2009 and the residential data is from 2015. Note that the dates in both years pertain to the same days of the week. 
The household and commercial demand data for Aug 3 were used to determine the set of houses included on the feeder and construct the plant.
To generate the MLR regression models of the AC and OL demand, we use data from June 24 to Aug 2, 2015 and commercial data from June 24 to Aug 2, 2009. The LTI and LTV models of the AC demand were constructed using device-level data from individual air conditioners from May 2 to Aug 2, 2015. The TOD regression models and Kalman filter covariance matrices were generated using data from the week preceding Aug 3. All testing and training data are available at \cite{ledva2017data}.

\subsection{Investigation of Algorithm Performance}
\label{sec:mainSimulations}

Table~\ref{table:dmdParams} gives the settings of $\eta^\text{s}$, and we set $\lambda = \eta^\text{r}=   1.0\times 10^{-5}$ across all simulations (with the exception of sensitivity simulations in Sections~\ref{sec:etaSweep} and \ref{sec:lambdaSweep}). Parameter $\lambda$ dictates the amount of weight shared amongst the models, where values near 1 force the DFS algorithm to generate estimates that are close to an average of the predictions of all models. By using a $\lambda$ value near 0, a single model can dominate the estimate if one model is more accurate than the rest. Parameters $\eta^\text{r}$ and $\eta^\text{s}$ were roughly tuned to achieve qualitative characteristics of fast switching between models without over-fitting. The optimal $\eta^\text{r}$ and $\eta^\text{s}$ for a given day will generally not be optimal across all days, and so tuning to achieve the desired qualitative characteristics is appropriate. In practice, $\eta^\text{r}$ and $\eta^\text{s}$ can be tuned based on recent historical data, and $\lambda$ can be tuned based on the historical accuracy of the models within the algorithm. An avenue for future research is to develop methods for online parameter tuning using real-time data.

\begin{table}[t]
\vspace{6pt}
\centering
\caption{Parameter $\eta^\text{s}$ Used in the DFS Scenarios in Section~\ref{sec:mainSimulations}} \label{table:dmdParams}
 \begin{tabular}{r c c c c c c}
\hline
Model Set & $\mathcal{M}^\text{Full}$ & $\mathcal{M}^\text{Full}$ & $\mathcal{M}^\text{Red}$ & $\mathcal{M}^\text{Red}$ & $\mathcal{M}^\text{KF}$ & $\mathcal{M}^\text{KF}$ \\
Update Method & 1 & 2 & 1 & 2 & 1 & 2 \\
\hline\hline
$\eta^\text{s}$ & 0.013 & 0.015 & 0.4 & 0.013 & 0.4 & 0.5 \\
\hline
\end{tabular}
\vspace{-10pt}
\end{table}

\begin{figure}
\centering
\subfloat[Total Demand]{
	\includegraphics[scale=1]{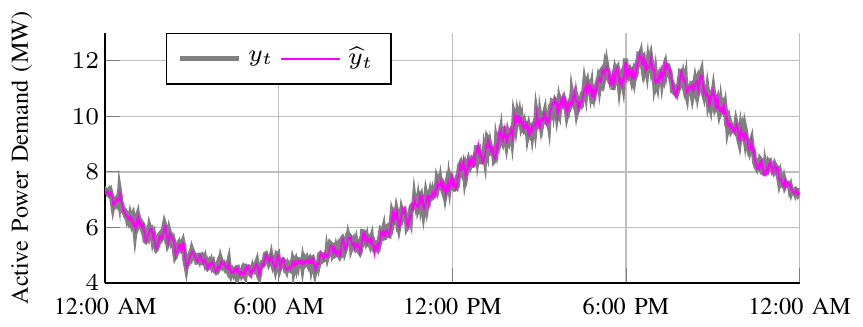}
	\label{fig:overallTimeSeries}
}
\\
\vspace{-15pt}	  
\subfloat[OL Demand]{
	\includegraphics[scale=1]{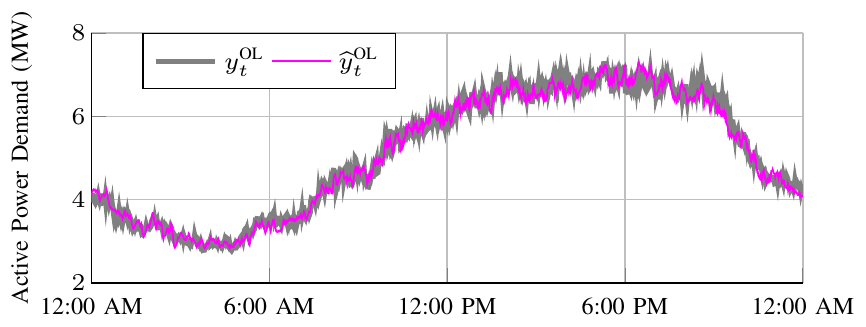}
	\label{fig:uncontrollableTimeSeries}
}
\\
\vspace{-15pt}	  
\subfloat[AC Demand]{
	\includegraphics[scale=1]{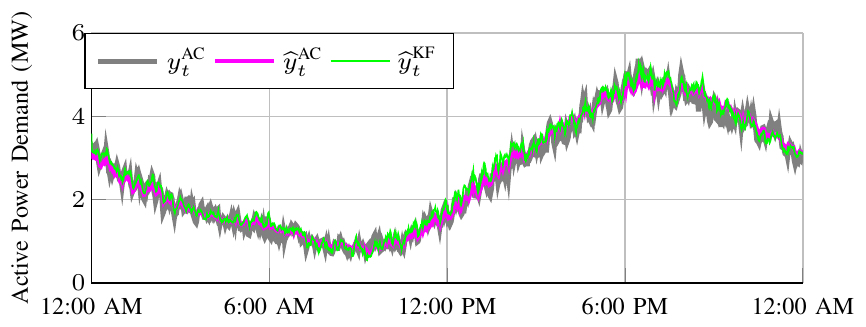}
}
\caption{Total, OL, and AC demands versus their DFS estimates (Aug 17, $\mathcal{M}^\text{Red}$, Update Method 1). The best Kalman filter estimate of the AC demand is also shown.} \label{fig:resultsTimeSeries}
\vspace{-10pt}	       
\end{figure}

\begin{figure}
    \centering
	\includegraphics[scale=1]{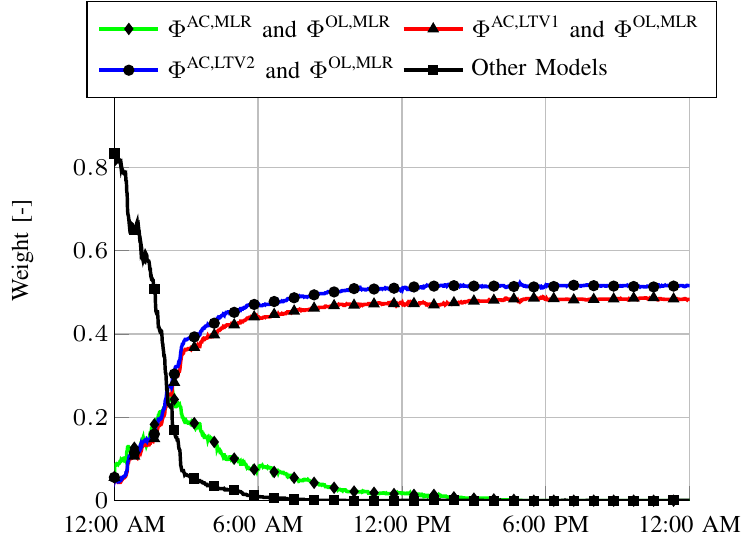}
	\vspace{-5pt}
	\caption{Model weights (Aug 17, $\mathcal{M}^\text{Red}$, Update Method 1).}
	\label{fig:weightTimeSeries}
\end{figure}

Figure~\ref{fig:resultsTimeSeries} depicts time series for the Aug 17 simulation with $\mathcal{M}^\text{Red}$ while using Update Method 1. 
Figure \ref{fig:weightTimeSeries} shows the evolution of the dominant model weights. The weights of the remaining models are summed and referred to as ``Other Models." In this scenario, the total demand is accurately separated into its AC demand and OL demand components in real-time, where the RMSE of the total demand, AC demand, and OL demand is $93.2$ kW, $151.0$ kW, and $150.8$ kW, respectively. In this scenario, DFS produces a more accurate AC demand estimate than BKF, which has an AC demand RMSE of $177.3$ kW. The RMSE of the AC demand for AKF is $214.0$ kW. The majority of the weight is initially given to the ``Other Models," because we initialize all models with the same weight. As the simulation progresses, the weight shifts between different model combinations. Since the combinations $\{\Phi^\text{AC,LTV2}, \Phi^\text{OL,MLR}\}$ and $\{\Phi^\text{AC,LTV1}, \Phi^\text{OL,MLR}\}$ perform best, they eventually earn more weight and dominate the predictions.   It is unsurprising that the $\Phi^\text{OL,MLR}$ is the most accurate OL demand model as it captures weather and time variables with the most detail, and it is unsurprising that $\Phi^\text{AC,LTV1}$ and $\Phi^\text{AC,LTV2}$ are the most accurate AC demand models as they capture the physical phenomenon driving changes in the AC demand as the outdoor temperature changes.

\begin{figure}
\centering
\subfloat[Update Method 1]{
	\includegraphics[scale=1]{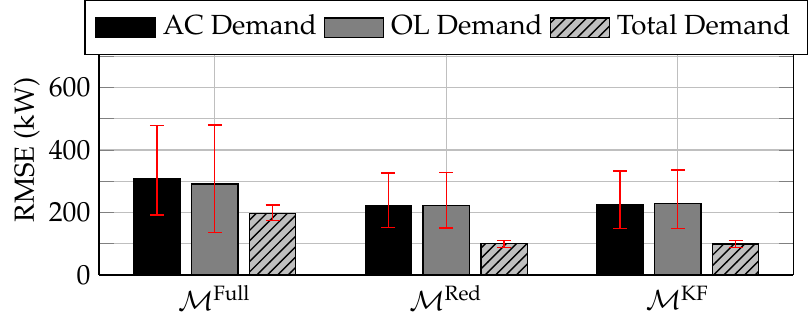}
	\label{fig:method1RmseRanges}
}
\\
\vspace{-5pt}	  
\subfloat[Update Method 2]{
	\includegraphics[scale=1]{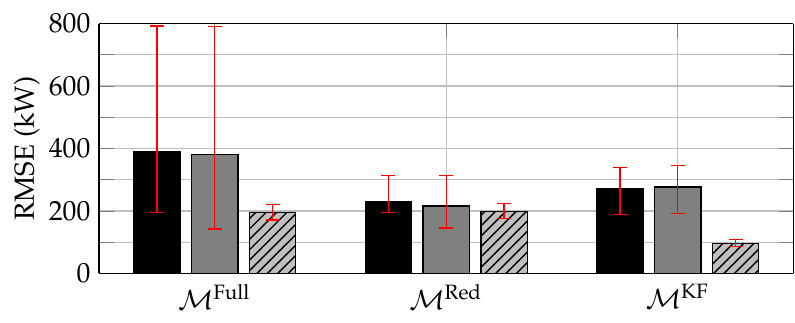}
\label{fig:method2RmseRanges}
}
\vspace{-5pt}	       
\caption{Minimum, mean, and maximum RMSE (kW) for the DFS scenarios in Section~\ref{sec:mainSimulations}.} \label{fig:rmseAndRanges}
\vspace{-10pt}	       
\end{figure}

Figure~\ref{fig:rmseAndRanges} presents the minimum, mean, and maximum RMSE across the full set of testing days for the total demand, the AC demand, and the OL demand for each DFS scenario. For comparison, BKF achieves a minimum RMSE of 148.4 kW, a mean RMSE of 195.3 kW, and a maximum RMSE of 318.9 kW for the AC demand, and AKF achieves a minimum RMSE of 173.1 kW, a mean RMSE of 259.4 kW, and a maximum RMSE of 357.5 kW for the AC demand. The model corresponding to the BKF varies from day to day and so it is not possible to obtain a single Kalman filter that always outperforms DFS.\footnotemark  To demonstrate the value of the measurement-based updates, we generated results for the full set of days using the model set $\mathcal{M}^\text{Red}$ and with $\eta^\text{s} = 0$; the measurement-based update is irrelevant with this parameter setting. The resulting total demand, AC demand, and OL demand RMSEs were $260.4$ kW, $254.2$ kW, and $245.2$ kW, respectively. These are significant increases over the DFS scenarios using $\mathcal{M}^\text{Red}$. 

\footnotetext{Choosing BKF on a particular day and applying it to all other days, we find that DFS performs better on approximately half of the days when using $\mathcal{M}^\text{Red}$. However, the loss function, divergence function, and initial model weights within DFS could be modified based on historical performance, which would improve its performance relative to the Kalman filter.}

The scenarios using $\mathcal{M}^\text{Full}$ have significantly higher AC demand RMSEs than the simulations using $\mathcal{M}^\text{Red}$ as well as the BKF and AKF simulations. Each of the LTI models may be accurate for a portion of the day when the AC demand is near the steady-state demand of the particular model. However, as the AC demand changes due to changes in the outdoor temperature, a given LTI model will become highly inaccurate. The DFS algorithm takes time to shift weight from the inaccurate model that was heavily weighted to the new model, and this results in increased RMSE. Eliminating these ``bad models'', by using $\mathcal{M}^\text{Red}$ rather than $\mathcal{M}^\text{Full}$, eliminates this issue.

The scenarios using $\mathcal{M}^\text{Red}$ generally do better, in terms of AC demand RMSE, than AKF and worse than BKF. On some simulated days DFS also outperforms BKF, as was shown in Fig.~\ref{fig:resultsTimeSeries}. Within this set of simulations, Update Method 2 results in higher AC demand RMSE than Update Method 1. The increased RMSE in Update Method 2 versus Update Method 1 can be explained due to the usage of only two discrete states within the LTV models. Specifically, the states reach their steady-state values rapidly, and so the measurement-based updates to the state do not persist for very long, whereas the measurement-based updates to the output used in Update Method 1 do. Using LTV models with more discrete states may allow Update Method 2 to achieve better RMSE, but this would complicate system identification. 

Finally, the scenarios with $\mathcal{M}^\text{KF}$  result in larger AC demand RMSE than those with $\mathcal{M}^\text{Red}$. The MLR model of the AC demand is often weighted heavily in the $\mathcal{M}^\text{Red}$ simulations, especially for Update Method 2. Given this, it makes sense that excluding this model would result in increased RMSE. 

\subsection{Sensitivity to the Parameters $\eta^\text{r}$ and $\eta^\text{s}$}
\label{sec:etaSweep}

We apply DFS to the full set of days while varying $\eta^\text{r}$ and $\eta^\text{s}$ to investigate the impact of those parameters on the results. We vary $\eta^\text{s}$ from 0.0 to 0.9 using increments of 0.1, and we vary $\eta^\text{r}$ from $10^{-7}$ to $10^{-3}$, where we increment the order (i.e., $10^{-7}$, $10^{-6},\dots$).  We apply DFS using every combination of these parameter values while using Update Method 1, $\mathcal{M}^\text{Red}$, and $\lambda = 1.0\times 10^{-5}$ as in Section~\ref{sec:mainSimulations}.

\begin{figure}[t!]
\centering
\includegraphics[scale=1]{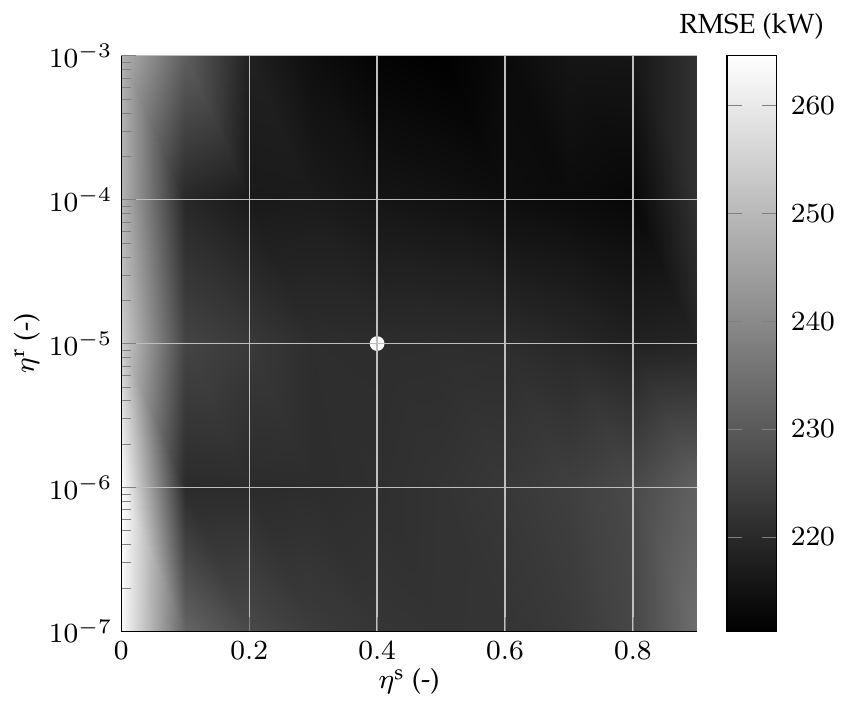}
\vspace*{-2pt}
 \caption{Average RMSE of the estimated AC demand across all days as a function of $\eta^\text{s}$ and $\eta^\text{r}$, using Update Method 1, $\mathcal{M}^\text{Red}$, and $ \lambda = 1.0\times 10^{-5}$ where the marker indicates the parameter values used in Section~\ref{sec:mainSimulations}. } \label{fig:etaSweeps}
\end{figure}

Figure~\ref{fig:etaSweeps} provides the average RMSE of the AC demand across the full set of testing days for each parameter value combination. With $\eta^\text{s}$ near zero the RMSE is relatively large as DFS makes small adjustments to the model predictions based on the realized prediction errors. The RMSE with $\eta^\text{s}$ near zero decreases slightly as $\eta^\text{r}$ increases because this allows for faster transitions in the weighting of the models. However, it should be noted that at larger $\eta^\text{r}$ values (e.g., $10^{-3}$), the model weights within DFS become erratic or noisy, and overfitting is possible. The RMSE is also relatively high with large $\eta^\text{s}$ (e.g., 0.9) and small $\eta^\text{r}$ as DFS adjusts the model predictions too aggressively and the model weights change slowly. Alternatively, as $\eta^\text{r}$ increases with large $\eta^\text{s}$, the RMSE decreases, but again the weights become vulnerable to overfitting. The RMSE using moderate $\eta^\text{s}$ values (e.g., from 0.2-0.7) are similar. The $\eta^\text{s}$ and $\eta^\text{r}$ values used within Section~\ref{sec:mainSimulations} do not achieve the lowest RMSE, but they achieve low RMSE while ensuring changes in the weights are reasonably fast but not erratic.  

\subsection{Sensitivity to the Parameter $\lambda$}
\label{sec:lambdaSweep}

We apply DFS to the full set of days while varying $\lambda$ from  $1.0\times 10^{-7}$ to $1.0$ to investigate the impact of $\lambda$ on the results. Within DFS, we use Update Method 1 and $\mathcal{M}^\text{Red}$, and we also set $\eta^\text{s} = 0.4$, and $\eta^\text{r} = 1.0\times 10^{-5}$ as in \ref{sec:mainSimulations}. 

\begin{figure}
    \centering
	\includegraphics[scale=1]{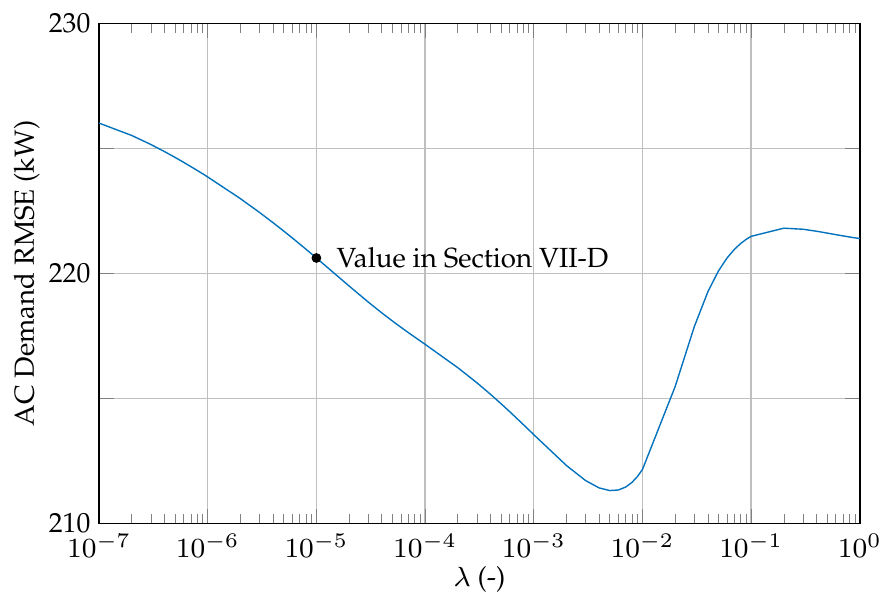}
	\vspace{-5pt}
	\caption{Average RMSE of the estimated AC demand across all days as a function of $\lambda$, using Update Method 1, $\mathcal{M}^\text{Red}$, $\eta^\text{s} = 0.4$, and $\eta^\text{r} = 1.0\times 10^{-5}$.}
	\label{fig:lambdaSweep}
\end{figure}

Figure~\ref{fig:lambdaSweep} gives the average RMSE of the estimated AC demand across all days as a function of $\lambda$. The average RMSE of the AC demand decreases as $\lambda$ is increased from $1.0\times 10^{-7}$ and reaches a minimum RMSE of 211.3 kW with $\lambda = 0.005$. As $\lambda$ increases from $0.005$, the RMSE increases until it remains relatively constant from $0.1$ to $1.0$. While we set $\lambda$ in Section~\ref{sec:mainSimulations} to allow a single model to dominate the DFS estimate if one model proved to be more accurate than the rest, Fig.~\ref{fig:lambdaSweep} indicates that tuning $\lambda$ on a set days that are similar to the testing days may allow a reduction in the RMSE.
 
\section{Conclusions}
\label{sec:conclusions}

In this paper, we applied an online learning algorithm, DFS, which uses DMD together with the Fixed Share algorithm, to estimate the real-time AC demand on a distribution feeder using feeder demand measurements,  weather data, and system models. Two implementations of algorithms based on DMD were developed and compared via case studies.  Our results showed that DFS can effectively estimate the real-time AC demand on a feeder. DFS achieved lower AC demand RMSE than the average across a set of Kalman filters. When selecting the most accurate Kalman filter \emph{ex post}, DFS generally results in larger RMSE. However, DFS learns the most accurate model, or combination of models, in real-time whereas the best Kalman filter can only be chosen after the simulation. The performance of DFS depends heavily on the inclusion of models within its set. Including models that are inaccurate for majority of the day degraded the algorithm performance as did removing models that were frequently weighted heavily. 

In this work, we separated the demand into only two components. However, the algorithm is applicable to scenarios with more than two components, assuming that we have at least one model of each demand component. As the number of components increases, it may become more difficult to disaggregate them, but these difficulties could be counteracted by incorporating more real-time measurements, e.g., the reactive power demand. Future work will develop improved AC demand models, investigate the relationship between the DMD and Kalman filter algorithms, and incorporate active control into the problem framework. 

\section*{Acknowledgments}
We thank the Pacific Gas \& Electric Company for the commercial building electric load data.

\bibliographystyle{./IEEEtran}
 \bibliography{./bibliographyFiles/nonDemandResponseCitations,./bibliographyFiles/demandResponseLitReview,./bibliographyFiles/energyDisaggregation}

\end{document}